# Knowledge Graphs and Knowledge Networks: The Story in Brief [1]

Amit Sheth, AI Institute, University of South Carolina
Swati Padhee and Amelie Gyrard, Kno.e.sis-Wright State University

## Abstract

*Knowledge Graphs (KGs) represent real-world noisy raw information in a structured form, capturing relationships between entities. However, for dynamic real-world applications such as social networks, recommender systems, computational biology, relational knowledge representation has emerged as a challenging research problem where there is a need to represent the changing nodes, attributes, and edges over time. The evolution of search engine responses to user queries in the last few years is partly because of the role of KGs such as Google KG. KGs are significantly contributing to various AI applications from link prediction, entity relations prediction, node classification to recommendation and question answering systems. This article is an attempt to summarize the journey of KG for AI.*

**Since 1735** when Leonhard Euler presented a solution to the Seven Bridges of Königsberg problem, graphs have emerged to graph databases, knowledge graphs, knowledge networks, social networks, and many more. The 1974 paper [2] may likely have given the first recorded definition of KG: "a mathematical structure with vertices as knowledge units connected by edges that represent the prerequisite relation". Knowledge Graphs: representation and structuring of scientific knowledge. A Ph.D. dissertation [3], which carried out a more systematic exploration described the KG as "a way of structuring and representing text encoding scientific knowledge". A particularly lovely review of the definitions of KGs appears in [4]. Given that KGs are one form of knowledge representation, it is natural to wonder about their relationship with other significant forms referred to as Semantic Networks, Conceptual Graphs, and Ontologies - the reader can find extensive literature on these concepts credited to John Sowa and Tom Gruber.

## What propels popularity of KG?

Primary reasons for the popularity of KGs in this century are: enabling new generation of applications for (prefacing with "semantic") search, browsing, recommendation, personalization, advertisement, etc. - both for the open Web as well as enterprises, and enhancing already very popular AI techniques of machine learning and natural language processing (NLP).

*KG enabled Web and Enterprise Applications:* With the rapid growth of the Web in the 1990s, the most important application was that of search. However, the search results still depended on the right selection of keywords. The first generation of commercial semantic applications that were empowered by KG or equivalent appeared around 2000. Taalee demonstrated web-scale semantic search, browsing, personalization and advertisement that was powered by large KG (called WorldModel or ontology in the patent (http://bit.ly/2FFu0gn)) with the RDF/S type schema and hundreds of millions of instances (triples with metadata) aggregated from many sources covering around 25 (sub) domains. However, KGs captured the broader mindset over a decade later when IBM's Watson won jeopardy against human experts in 2011 and Google semantic search rolled out in 2012. While these systems used machine learning (ML) and natural language processing (NLP), these systems demonstrated the indispensable role of KG (for a perspective: bit.ly/15yrSemS) in critical areas of computing. The personal assistant and smart speakers such as Amazon Alexa are also heavily reliant on KG, so their rapid growth (recently estimated at 78% year-over-year) reinforce the importance of KGs. There are still limitations when we are not querying the right keyphrase, as a human, we need to formulate the query to handle synonyms to get more appropriate results. For instance, "knowledge-based", "semantic-based", "ontology-based" are synonyms of KGs that must be considered when investigating the topic. Interacting with the Web this way aggregates additional results and avoid neglecting the domain knowledge already specified within past projects.

Today, tech giants including Microsoft, Siemens, LinkedIn, Airbnb, eBay, and Apple, as well as smaller companies (e.g. ezDI, Fraanz, Metaphactory/Metaphacts GmbH, Semantic Web Company GmbH,

---

1. This is authors' preprint version for (cite as): A. Sheth, S. Padhee and A. Gyrard, "Knowledge Graphs and Knowledge Networks: The Story in Brief," in IEEE Internet Computing, vol. 23, no. 4, July-Aug. 2019.

Mondeca, Stardog, Diffbot, Siren) are using Enterprise KGs (which are often proprietary but may incorporate public knowledge such as DBPedia) and KG-enabled technologies for critical products and services for its customers (e.g., Maana). The importance of proprietary KGs for some companies has justified the use of thousands of employees supporting technical and editorial (curation) activities (bit.ly/KG10k).

*__Enhancing learning__*: AI techniques including ML algorithms which learn from pre-labeled examples are acknowledging that "data alone is not enough" [5]. There is a growing body of work seeking to demonstrate how and how much use of domain knowledge improves the results or effectiveness of state of the art ML and NLP techniques [6]. Learning the underlying patterns in the data goes beyond instance-based generalization to some external knowledge represented in structured graphs or networks. Deep Learning (DL) has shown significant advances in improving NLP by probabilistically learning latent patterns in the data using a multi-layered network of computational nodes (i.e. neurons/hidden units). However, with the tremendous amount of training data, uncertainty in generalization on domain-specific tasks, and delta improvement with an increase in complexity of models seem to raise a concern on the features learned by the model. Utilization of prior knowledge will aid in supervising the learning of features and brings in explainability. The next opportunity could be to complement the implicit or later knowledge (entities and relationship) by KGs that already capture synonyms and variants of entities and a variety of typed relationships. Many challenges remain, such as how to represent the knowledge propagation between nodes as complex real-world relationships in a graph. Pioneers in AI are hence manipulating the structured KGs for DL with relational inductive biases (zd.net/2Jblg2A), transfer learning (inter-domain knowledge sharing) and other new methods of infusing KG into ML. Although much work remains, we think KGs will play an increasing role in developing hybrid neuro-symbolic systems (that is bottom up deep learning with top down symbolic computing) as well as in building explainable AI systems for which KGs will provide a scaffolding for punctuating neural computing.

> **Knowledge Graph (KG)** is a structured knowledge in a graphical representation. KG can be used for a variety of information processing and management tasks such as:
> 1) enhanced (semantic) applications such as search, browsing, personalization, recommendation, advertisement, and summarization
> 2) improving integration of data, including data of diverse modalities and from diverse sources,
> 3) empowering ML and NLP techniques, and 4) improve automation and support intelligent human-like behavior and activities that may involve conversations or question-answering and robots.

> **Knowledge Networks (KN)** integrate and combine knowledge (usually captured as KGs) from various domains. Knowledge networks should have schemas, datasets, and documentation to explain their usability across applications, and provide "horizontal services" to support knowledge-intensive applications, and may specialize to focus on a chosen domain (i.e., "vertical KN", as in neuroscience KN), and interconnect multiple fields to create a cross-domain KN.

In the remainder of this article, we review the development of KGs, ambitious use cases, emerging research challenges, and discuss the emergence of Open Knowledge Networks to conclude the article.

## Development of KGs

KGs are curated through manual, semi-automatic, or automatic approaches. These approaches support extraction from semi-structured and structured data (DBpedia, YAGO), unstructured data (NELL), HTML web pages, books, and Microdata annotations on the Web (Google's Knowledge Vault), public collaborative data like Wikipedia and Freebase (Yahoo's KG), collaborative manual editors (Wikidata), etc. When the source of knowledge itself is not high quality, as is the case for the majority of situations, curation is crucial to ensure the quality of the KGs and ultimately, their usability. We review some of the significant knowledge sources and KG creation efforts. Some of the endeavors in KG curation include:

*__Linked Open Data__* (LOD) provides diverse sources of knowledge to populate and enrich KGs. By March 2019, it covered 1,239 datasets with 16,147 links. In spite of its vast size, breadth of coverage and quality (timely update, provenance, context), LOD has its inherent challenges in terms of curation, usability, temporal validity of datasets, missing domains and more. Recently, Google Dataset Search (toolbox.google.com/datasetsearch) started to provide a friendly user interface to crawl, search, and query existing datasets accessible on the Web, similar to LOD.

*__Schema.org__* (schema.org/) demonstrates the impact and need to structure and interlink data on the Web. Schema.org was created by major search engine companies such as Bing, Google, and Yahoo, in an agreement with the community that designs vocabularies to annotate Web pages. Annotations are embedded in websites to provide structure to data on the Web (e.g., restaurant opening hours) that are frequently searched. Its success is also partly due to the adoption by Content Management Systems (CMS) such as Drupal which automatically annotates web pages by referencing to schema.org classes and properties. The easy-to-use examples (without markup, Microdata, RDFa, and JSON-LD) provided by the Schema.org documentation encourages dissemination of new technologies. As a result, a growing number of corporations are now adopting KGs by agreeing to use a standard vocabulary.

*__Data Commons Knowledge Graph__* (DCKG) (datacommons.org/) designs "data as a service" approach with a simple browser and API. This "data as a service" approach eases the tedious and cumbersome task of dealing with datasets (e.g., download, integration), using the schema.org vocabulary to aggregate data from Wikipedia, the US Census, NOAA, and FBI. It unifies the way to describe cities, counties, states, countries, congressional districts, census estimates, labor statistics, crime data, health data, biological specimens, power plants, and ENCODE (Encyclopedia odd DNA elements). Furthermore, the provenance of the dataset is explicitly described.

*__Wikidata__* (wikidata.org/) is Wikipedia's open-source machine-readable database with millions of entities where everyone can contribute and use (with reading and editing permissions) with a user-friendly query interface. It covers a wide variety of domains and contains not only textual knowledge but also images, geo-coordinates, and numerics. Wikidata uses unique identifiers for each entity/relation for accurate querying and provides provenance metadata, unlike DBpedia and Schema.org. For instance, it includes information about a fact's correctness in terms of its origin, temporal validity (reference point of time during of the fact). Wikidata is one of the latest projects acknowledging the dynamic nature of KG and is continuously updated by human contributors unlike DBpedia which is curated from Wikipedia once in a while.

These vast and open knowledge repositories serve as ecosystems to unify data management, interoperability, innovation, and entrepreneurship across different domains enhancing AI applications in agriculture, trust, health, real-time situational-awareness, and more. In the next section, we demonstrate the need to interlink domains to build innovative AI applications.

**Promising Cross-Domain Use Cases Demonstrating KG Impact**

Finding, reusing, and interlinking the cross-domain knowledge (knowledge from multi-disciplinary fields) described in the previous section are some of the future challenges requiring domain-specific knowledge extraction. For instance, temporal analysis of events will rely on the temporal evolution of event-specific knowledge (e.g., via disease-specific health KG) and personalized analysis of data (e.g., via patient-specific health KG). Hereafter we briefly illustrate some emerging cross-domain use cases that demonstrate the indispensable role of KGs:

*__Use case 1:__* Designing a cooking robot requires cross-domain knowledge such as: 1) observational (sensory data) and common-sense knowledge to perceive the surrounding environment, 2) knowledge representation to model the knowledge concerning the surrounding environment, 3) appropriate cross-domain knowledge reasoning mechanisms, and 4) services for end-users with a friendly interface (GUI or chatbots). RoboBrain (www.technologyreview.com/s/533471/) is one of such efforts to model all the multimodal cross-domain knowledge required by a robot to be smarter in every field, including the kitchen.

*Use case 2:* The field of cognitive science covers a broad scope of human intelligence, including linguistics, mathematical-logical, musical, bodily-kinesthetic, social (interactions and relationships), emotional (empathetic and moral), and personal knowledge [Fig.1]. For instance, to "inject" human intelligence into AI assistants such as Amazon Alexa, utilization of cross-domain knowledge of social interactions, emotions and linguistic variations of natural language is critical.

*Use case 3:* Researchers are trying to model empathy and morality into self-driving cars. For instance, Morale machine platform by MIT (moralmachine.mit.edu/) gathers human knowledge on moral decisions to model machine intelligence for self-driving cars. For AI agents to mimic human emotions and decisions, we need to model human emotional knowledge of empathy, moral, and ethics.

*Use case 4:* IBM Watson, Assistant for Health Benefits, is now capable of personalizing interactions with its members. Smart health agents are adapting to answer real-world personalized complex health queries in simple interactive language. Developing these healthcare chatbots requires patients' environmental knowledge, health data, and coordination with their healthcare physicians. Researchers are developing several knowledge-enhanced chatbots for healthcare (bit.ly/Hcbots) e.g., asthma (bit.ly/kBot), depression (bit.ly/ReaCTrack), and obesity.

## Emerging Challenges in KG

The use cases presented above present new challenges to researchers such as the following:

*Challenge 1: Capturing Context*

In this era of AI-infused systems for applications including conversational virtual agents, and the advancements in Human-Computer-Interactions (HCI), context is the key for more sensible conversations. While researchers are trying to capture context in algorithms using Reinforcement Learning, KGs are emerging with contextual reasoning (formalization, representation and standardization of provenance, time, location, uncertainty, evidence), contextualized inference rules including schema integration, private-public data sharing policies and authorizations, query syntaxes for scalability, feasibility for end-users, etc. Such representations are required to ascertain spatiotemporal validity of facts as well.

*Challenge 2: Domain-Specific Knowledge Extraction*

We have discussed the impact of KG for various applications. However, domain-specific knowledge (real-time as well as background knowledge) is critical for extraction of task-specific assertions and their normalization, as general KGs such as DBpedia do not provide the application-specific knowledge necessary for effective and efficient reasoning. Recent studies in domain-specific subgraph extraction have significantly contributed to improving the efficiency and quality of information extraction and complex task-specific algorithms by capturing contextually relevant factual knowledge [7].

*Challenge 3: Knowledge Alignment*

With multiple approaches, data sources, and technologies for KG curation, interlinking of KGs faces the challenge in the alignment of underlying knowledge representation. For example, an object "apple" can be represented as a concept or an instance depending on the underlying schema and representation. NLP and ML/DL techniques are being used on a set of schemas to extract, understand, and summarize the structured knowledge encoded in a processable machine format (e.g., transfer learning for taxonomy or schema alignment).

*Challenge 4: Real-Time KG for Fast Data*

We are facing a new challenge with the birth of Fast data i.e., real-time data or streaming data for quick decisions. Today, many industries are relying on fast data analysis solutions for near real-time or streaming multimodal data. While KG for Big Data has already gained the attention of the research community, real-time building KG from Fast Data with agility and efficiency is an emerging issue. IBM Fast Data Platform is one of the initial efforts with real-time data and ML algorithms to cater to streaming applications.

*Challenge 5: Quality and Validity of KGs*

Knowledge extraction approaches vary from manual to semi-automatic to automatic techniques, including statistical methods such as Relational ML for predicting new facts and edges [8]. Either way, errors, and missing knowledge can quickly proliferate, leaving the knowledge incomplete or incorrect. Knowledge refinement for adding such missing knowledge or identifying and correcting the erroneous knowledge using holistic and automatic approaches to improve KG quality is a thriving research area [9]. The correctness of knowledge added to KGs also depend on its temporal validity. Temporally changing relationships to define the relatedness between entities to model domain-specific [10] and temporal multi-relational data is another concern deserving significant attention.

*<u>Challenge 6: Adaptive Knowledge Network</u>*

Change is a law of nature, and static KGs like DBpedia fail to capture this dynamic flow of information. Diverse applications of AI are increasingly relying on the knowledge which is not necessarily static e.g., President_of_the_USA, champion_of_FIFA_World_Cup are temporally-sensitive facts, unlike birth_date or death_date. The need for accurate temporal query responses by dominant search engines requires extracting, maintaining, and updating the temporal facts in KGs. Analyzing real-world dynamic events (e.g., elections, natural disasters, etc.) requires real-time predictive analysis, trend analysis, spatiotemporal decision making, public opinion analysis. Researchers have started to curate Adaptive Knowledge Networks from incoming real-time multimodal spatiotemporally evolving data which evolve with time [11].

## Open Knowledge Networks

At the heart of the above challenges lies this one big question "Can we access all of these KGs and put them to use?". AI systems like a cooking robot or Ambient Assisted Living (AAL) require multimodal knowledge from multiple domains and sources. For instance, monitoring of elderly needs knowledge of their biometrics, home-conditions, disease history, and real-time behavior (fall detection) in addition to biomedical and common-sense knowledge. These applications need a knowledge network (KN) which interlinks multimodal cross-domain knowledge curated from various sources as well as a personalized KG for healthcare.

However, many of the KGs in this KN are proprietary and expensive for usage by academia or industry researchers and small clients. Recently, pioneers from Federal, industry, and academia have proposed an Open Knowledge Network (OKN) [12] to provide a nation-scale open infrastructure linking cross-domain information of relevant entities. OKN initiative is expected to advance the perception of KGs in AI to shape an open KG of all world knowledge represented as entities and relationships. It is anticipated to address a coverage of "macro (e.g., have there been unusual clusters of earthquakes in the US in the past six months?) to the micro (e.g., what is the best combination of chemotherapeutic drugs for a 56 y/o female with stage 3 brain cancer.)" [12]. OKN will encourage the development of query and fusion servers that integrates reasoning capabilities. It is proposed as one of the ten Big Ideas by the National Science Foundation (NSF) for solving complex problems by consolidating knowledge, tools, and expertise from multiple disciplines and forming novel frameworks to catalyze scientific innovation and discovery.

With OKN as an open and inclusive community, innovative applications in AI will prompt curation of reliable knowledge graphs/networks. NSF's OKN initiative which is part of Harnessing the Data Revolution Big Idea (www.nsf.gov/pubs/2019/nsf19050/nsf19050.jsp) as a follow up to the OKN series of the workshop has resulted in several soon to start efforts dealing with horizontal and vertical challenges are likely to accelerate progress in this area.

## Looking Forward to KG enabled AI systems that are more human-like

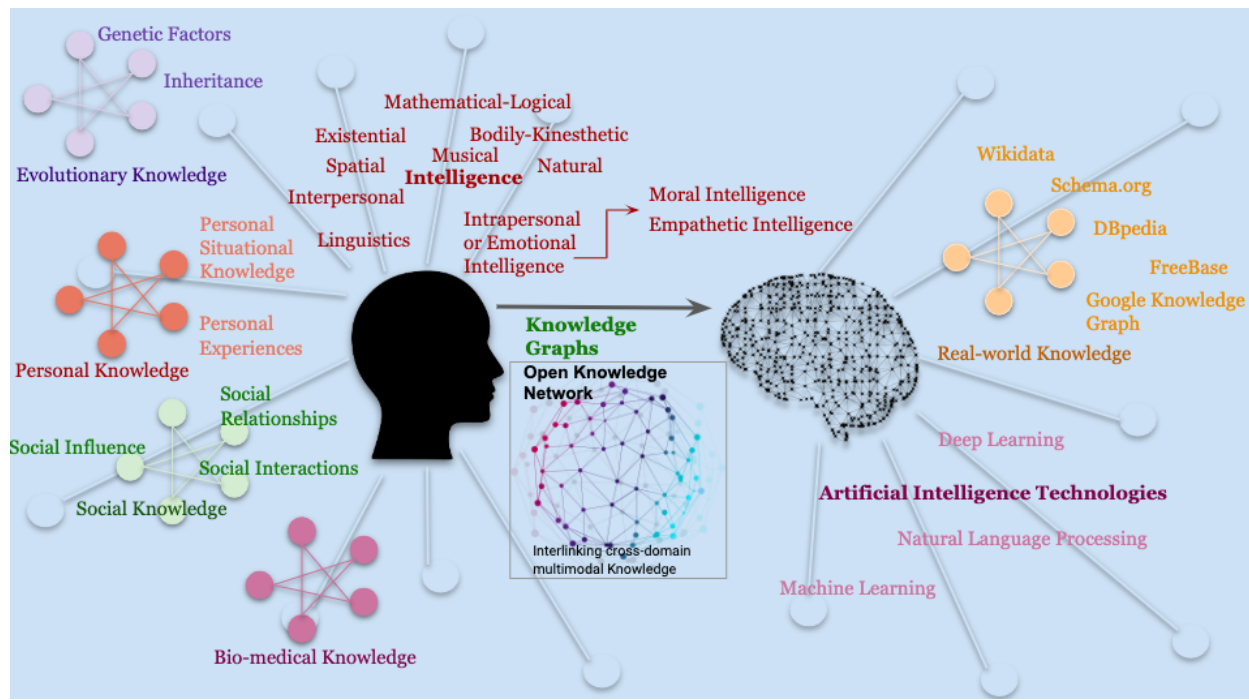

**Fig.1. An expanding role of Knowledge in future AI systems**

AI has evolved from the ancient Greek legends of golden robots to Sophia, the talking robot. Today, we are surrounded by AI systems with rapidly growing cognitive abilities. However, to facilitate more human-like machines, AI needs to mimic various aspects of human intelligence including developing a multifaceted understanding of all types of sensory and social data, and multimodal information [Fig.1] which integrates the ability to reason, perceive and learn from experiences, interactions, and surroundings. Although Sophia is capable of displaying some facial expressions, human emotions go beyond facial muscles (amusement, contempt, contentment, embarrassment, excitement, guilt, pride in achievement, relief, satisfaction, sensory pleasure, and shame [13]). While AI is attempting to mimic our inheritance with generative evolutionary algorithms, to be more human-like, it needs to infuse these sophisticated world knowledge and human emotions in formalized representations to develop the next-generation AI techniques. Only then, AI systems can differentiate between the tears of joy and the tears of sorrow and empathize with us interacting more humanely. Knowledge graphs and networks are likely to provide the underlying infrastructure on which advanced techniques will be built to progress towards this direction.

The purpose of this first article in this new department on KG is to give a quick overview of the concepts and phenomena of KG which have become indispensable to a growing number of AI techniques and solutions. Future articles will address a wide variety of topics and techniques associated with the emerging KG ecosystem.

**Amit Sheth** is the founding director of the AI Institute at the University of South Carolina. He is a fellow of IEEE, AAAI and AAAS. https://www.linkedin.com/in/amitsheth/,  amit@sc.edu

**Swati Padhee** is a PhD student working on dynamic knowledge graphs and its use to support temporal queries. swatipadhee.com, swati@knoesis.org

**Amelie Gyrard** is a postdoc with research interests in IoT and Semantic Web. amelie@knoesis.org